\documentclass[sigconf]{acmart}
\usepackage{bbding}
\usepackage{amsmath,amsfonts}
\usepackage{algorithmic}
\usepackage{algorithm}
\usepackage{array}
\usepackage[caption=false,font=normalsize,labelfont=sf,textfont=sf]{subfig}
\usepackage{url}

\usepackage{hyperref}
\hypersetup{pdfborder={0 0 0}, colorlinks=true, linkcolor=red, citecolor=blue}
\usepackage{textcomp}
\usepackage{stfloats}
\usepackage{url}
\usepackage{verbatim}
\usepackage{graphicx}
\usepackage{multirow}
\usepackage{color}
\usepackage{colortbl}
\definecolor{Ocean}{RGB}{129,154,254}
\definecolor{orange}{RGB}{254,144,95}
\definecolor{low}{HTML}{D6EAF8}
\definecolor{high}{HTML}{FADBD8}
\usepackage[utf8]{inputenc}
\usepackage{bm}
\usepackage{bbding}
\usepackage{subfig}
\usepackage{booktabs}
\usepackage{balance}
\usepackage[explicit]{titlesec}
 
\titlespacing*{\section}{0pt}{1.2ex plus .0ex minus .0ex}{.3ex plus .0ex}
\titlespacing*{\subsection}{0pt}{1.2ex plus .0ex minus .0ex}{.3ex plus .0ex}
\AtBeginDocument{%
  \providecommand\BibTeX{{%
    \normalfont B\kern-0.5em{\scshape i\kern-0.25em b}\kern-0.8em\TeX}}}

\setcopyright{acmlicensed}
\copyrightyear{2024}
\acmYear{2024}
\setcopyright{acmlicensed}\acmConference[MM '24]{Proceedings of the 32nd ACM International Conference on Multimedia}{October 28-November 1, 2024}{Melbourne, VIC, Australia}
\acmBooktitle{Proceedings of the 32nd ACM International Conference on Multimedia (MM '24), October 28-November 1, 2024, Melbourne, VIC, Australia}
\acmDOI{10.1145/3664647.3681168}
\acmISBN{979-8-4007-0686-8/24/10}

\begin{document}

\title{Digging into Contrastive Learning for Robust Depth Estimation with Diffusion Models}
\author{Jiyuan Wang}
\orcid{0009-0001-2525-7473}
\affiliation{%
  \institution{Beijing Jiaotong University}
  \city{Beijing}
  \country{China}}
\email{wangjiyuan9@163.com}

\author{Chunyu Lin*}
\orcid{0000-0003-2847-0349}
\affiliation{%
  \institution{Beijing Jiaotong University}
  \city{Beijing}
  \country{China}}
\email{cylin@bjtu.edu.cn}

\author{Lang Nie}
\orcid{0000-0002-7776-889X}
\affiliation{%
  \institution{Beijing Jiaotong University}
  \city{Beijing}
  \country{China}}
\email{nielang@bjtu.edu.cn}

 \author{Kang Liao}
 \orcid{0000-0001-9429-1096}
\affiliation{%
 \institution{Nanyang Technological University}
 \city{Singapore}
 \country{Singapore}}
\email{kang.liao@ntu.edu.sg}

 \author{Shuwei Shao}
 \orcid{0000-0001-8057-1599}
\affiliation{%
 \institution{Beihang University}
 \city{Beijing}
 \country{China}}
\email{swshao@baau.edu.cn}

  \author{Yao Zhao}
  \orcid{0000-0002-8581-9554}
\affiliation{%
 \institution{Beijing Jiaotong University}
 \city{Beijing}
 \country{China}}
\email{yzhao@bjtu.edu.cn}

\renewcommand{\shortauthors}{Jiyuan Wang et al.}

\begin{abstract}
Recently, diffusion-based depth estimation methods have drawn widespread attention due to their elegant denoising patterns and promising performance. 
However, they are typically unreliable under adverse conditions prevalent in real-world scenarios, such as rainy, snowy, etc. 
In this paper, we propose a novel robust depth estimation method called D4RD, featuring a custom contrastive learning mode tailored for diffusion models to mitigate performance degradation in complex environments. Concretely, we integrate the strength of knowledge distillation into contrastive learning, building the `trinity' contrastive scheme. This scheme utilizes the sampled noise of the forward diffusion process as a natural reference, guiding the predicted noise in diverse scenes toward a more stable and precise optimum. Moreover, we extend noise-level trinity to encompass more generic feature and image levels, establishing a multi-level contrast to distribute the burden of robust perception across the overall network. Before addressing complex scenarios, we enhance the stability of the baseline diffusion model with three straightforward yet effective improvements, which facilitate convergence and remove depth outliers.
Extensive experiments demonstrate that D4RD surpasses existing state-of-the-art solutions on synthetic corruption datasets and real-world weather conditions. Source code and data are available at \url{https://github.com/wangjiyuan9/D4RD}.
\end{abstract}



\maketitle

\begin{figure}[t]
\centering
\hspace{-5mm}
\includegraphics[width=\linewidth]{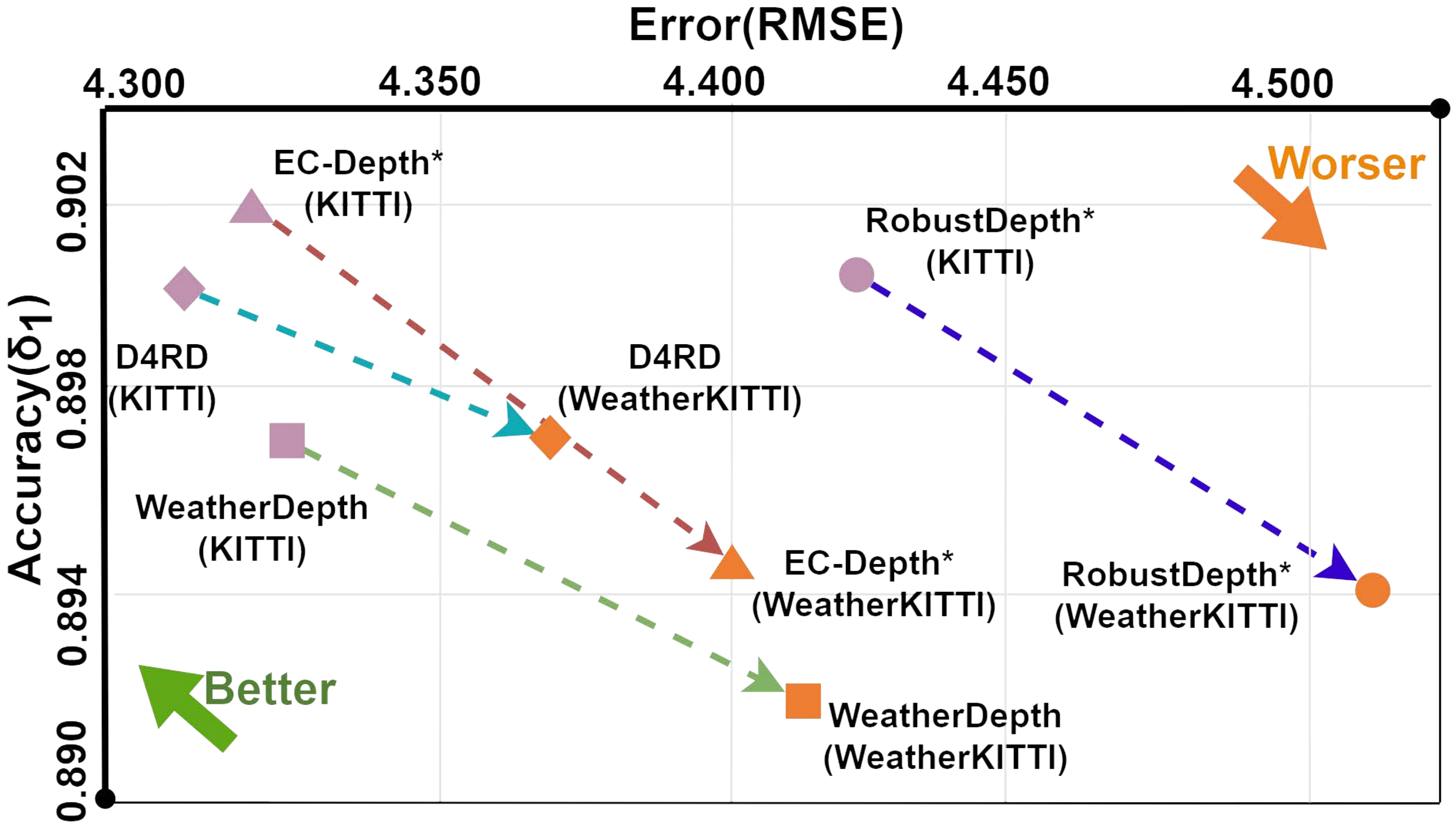}
\vspace{-0.3cm}
\caption{\label{fir} Comparisons with RMDE method\cite{wjy,iccv,ecdp} on KITTI and WeatherKITTI. The length of the Line represents the performance degradation magnitude under adverse environments. All the methods are trained on the same dataset WeatherKITTI for fairness.}
\vspace{-0.7cm}
\end{figure}

\section{Introduction}
Estimating accurate depth from a single image is highly attractive because this task arises whenever the 3D structure is needed. Since the supervised Monocular Depth Estimation (MDE) methods\cite{midas, dpt} require high-cost ground truth (GT) depth labels. To address this limitation, researchers have explored self-supervised approaches that leverage adjacent frame pose information and photometric consistency from video sequences. However, most existing self-supervised MDE models can only handle clear and ideal settings but are highly unreliable under challenging conditions. This limitation significantly hampers the practical applicability of these methods.

Recently, diffusion models have demonstrated their superiority in MDE\cite{cvprdfs,cvprdfs2} with remarkable performance. Moreover, some works from other domains\cite{otherrobust1, otherrobust2} have revealed that the denoising diffusion process allows the potential of robust perception. These findings have inspired us to explore how diffusion models can address the challenges faced in MDE and enhance robustness. 
To this end, we propose \textbf{D}iffusion for(\textbf{4}) \textbf{R}obust \textbf{D}epth \textbf{(D4RD)}, a novel diffusion-based MDE framework. D4RD incorporates a new multi-level `trinity' contrastive scheme designed to enhance robustness and mitigate performance degradation in challenging conditions. 

To better understand the concept of the `trinity', we need to revisit the previous efforts aimed at resolving the challenges of Robust Monocular Depth Estimation (RMDE). Generally, these approaches fall into two main categories: contrastive learning-based alignment approaches\cite{ecdp, allday, zcq, wjy} and knowledge distillation-based pseudo-supervised methods\cite{iccv,iccv2,eccv}. Considering that corruption does not affect the physical scene depth, the former approach typically involves the clear image along with its various augmentations into the network, enforcing consistency among different output depths to provide extra supervision. However, a common issue of these methods is that directly maximizing the depth similarity can easily fall into a collapsing solution\cite{simsiam}. To address this issue, additional guidance is required, which poses a challenge in a self-supervised manner since the photometric consistency cannot provide a perfect convergence anchor.

The latter usually employs a pre-trained teacher model to estimate the depths of clear images and uses these estimates as pseudo-labels to train the student model on augmented images. Obviously, there is a performance upper bound for the student model as the teacher model is not entirely accurate.

In short, we urgently need a perfect label that can not only serve as an anchor for guiding optimization in contrastive learning but also eliminate performance bottlenecks in knowledge distillation. This label seems impossible in self-supervised learning but we ingeniously find it in the diffusion process—the sampled noise. In the forward diffusion process, we obtain noisy images by adding a sampled noise that follows a normal distribution to the original image. During the training process, sampled noise is considered as the GT label for supervising noise prediction. When the prediction is accurate, the original image will be accurately restored\cite{ddpm, ddim}. Therefore, we incorporate this noise as supplementary guidance for contrastive learning to establish a `trinity' contrast pattern. Concretely, we maintain the original consistency constraints unchanged, but on this basis, we apply sampled noise constraints to clear/augment input noise predictions, gathering the pair of estimated noise towards a more stable and precise optimum. If the model is completely robust and accurate, then the three types of noise(\textit{i.e.}, clear/augmented/sampled noise) should be identical, that is, the `trinity'.

Moreover, \cite{forget} have found that the burden of robust dense perception is mainly concentrated on a specific component of the network (\textit{e.g.}, the lower layers), which limits the network potential. Based on this, we intuitively try to distribute these responsibilities across the network by extending the `trinity' paradigm to more generic feature and image levels. Naturally, finding suitable reference points like sampled noise is challenging on these occasions, so we rely on clear scenes and teacher models to provide guidance to the model. Despite that, we found our trinity is still better than the traditional distillation and contrast methods. In addition, to further explore how multi-level contrastive learning promotes network performance, we visualize the perceptual results after each level. The improvement, as depicted in Fig. \ref{mult-vis}, is likely attributed to each level addressing distinct aspects of image enhancement, ultimately leading to more robust and accurate depth results.

D4RD takes MonoDiffusion \cite{monodfs} as the baseline framework. Before incorporating the multi-level trinity contrastive paradigm, we also enhance the stability of the baseline with three simple but effective improvements. It strengthens convergence and eliminates depth outliers, which lay the foundation for handling complex scenarios. 
Finally, extensive experiments show that our proposed method, D4RD, achieves SoTA performance among all RMDE competitors across \textbf{two} syntheses dataset and \textbf{five} real weather conditions. Moreover, as shown in Fig. \ref{fir}, D4RD not only surpasses other methods with a significant margin but also demonstrates the minimum performance degradation in challenging conditions. In summary, the contributions of this paper are as follows:
\begin{itemize}
\item We enhance the diffusion-based depth estimation architecture with better stability and convergence, and further establish a novel robust depth estimation framework, named \textbf{D4RD}, with a customized contrastive learning scheme for diffusion models.

\item Benefiting from the natural guidance of sampled noise, we ingeniously integrate the strength of distillation learning into contrastive learning to form a `trinity' contrast pattern, which facilitates noise prediction accuracy and robustness in D4RD.

\item The noise-level trinity is then expanded to more generic feature and image levels, building the multi-level trinity scheme. It evenly distributes the pressure of handling domain variances across different model components.
\end{itemize}

\section{Related Work}

\subsection{Monocular Depth Estimation}
MDE aims to predict the accurate depth from a single image, which is an ill-posed problem. Intuitively, this challenge was initially seen as a dense regression task\cite{eigen, eigen2} based on GT depth labels. Four years later, Fu et al.\cite{fu} discovered that reconstructing MDE into a depth bins classification task can significantly improve performance. Still about four years later, as the diffusion model showcases its talents in generative task\cite{ddpm,ddim}, detection\cite{detect}, and segmentation\cite{segrefiner}, DDP\cite{DDP} first reformulates this task as a depth map denoising task, and lead to giant progress again. Followers like Diffusiondepth\cite{dfsdepth}, DDVM\cite{ddvm}, VPD\cite{vpd}, TAPD\cite{tapd}, EcoDepth\cite{cvprdfs2}, Marigold\cite{cvprdfs}, and MonoDiffusion\cite{monodfs} all demonstrate the advantages of this paradigm in various MDE sub-tasks. D4RD also adopts this promising scheme and extends it on robust depth estimation.

\subsection{Robust Depth Estimation}
As mentioned before, although MDE is an ill-posed problem, with the development of research and task optimization in the past decade, MDE on clear data sets has achieved excellent performance. Therefore, for RMDE tasks, an intuitive and effective method is to use the results of clear scenes, which can be divided into two major categories, knowledge distillation and contrastive learning. 

Distillation-based methods focus on leveraging the well-predicted depth map (typically estimated from clear scenes) as pseudo-labels to assist in training the RMDE model. MD4all\cite{iccv2} trained a base model on sunny scenes first, and then employed it as a teacher to train another network for weather scenarios with a specifically designed loss function. SSD\cite{eccv} may be the most similar work to ours because it also takes diffusion to solve RMDE tasks. However, it follows the distillation paradigm of MD4all and introduces an enhanced dataset generated by diffusion, even altering the scene itself when rendering weather or night enhancements. 

Contrastive-based methods typically improve the model's robustness by ensuring consistency between the depth predictions of a clear image and its augmented versions. Robust-Depth\cite{iccv} introduces the semi-augmented warp and bi-directional contrast, which ensures the consistency assumption and improves the accuracy. WeatherDepth considers the domain gap between clear scenes and complex scenes and solves this problem by applying a gradual adaptation scheme based on curriculum contrastive learning. Additionally, EC-Depth\cite{ecdp} combines contrast and distillation with double-stage training. The first stage uses contrast, creating a KL divergence-like contrast to align the depth of harsh scenes, and the second stage uses distillation, introducing EMA for teacher-student joint learning.

\section{Method}
\begin{figure*}[ht]
\centering
\includegraphics[width=0.95\textwidth]{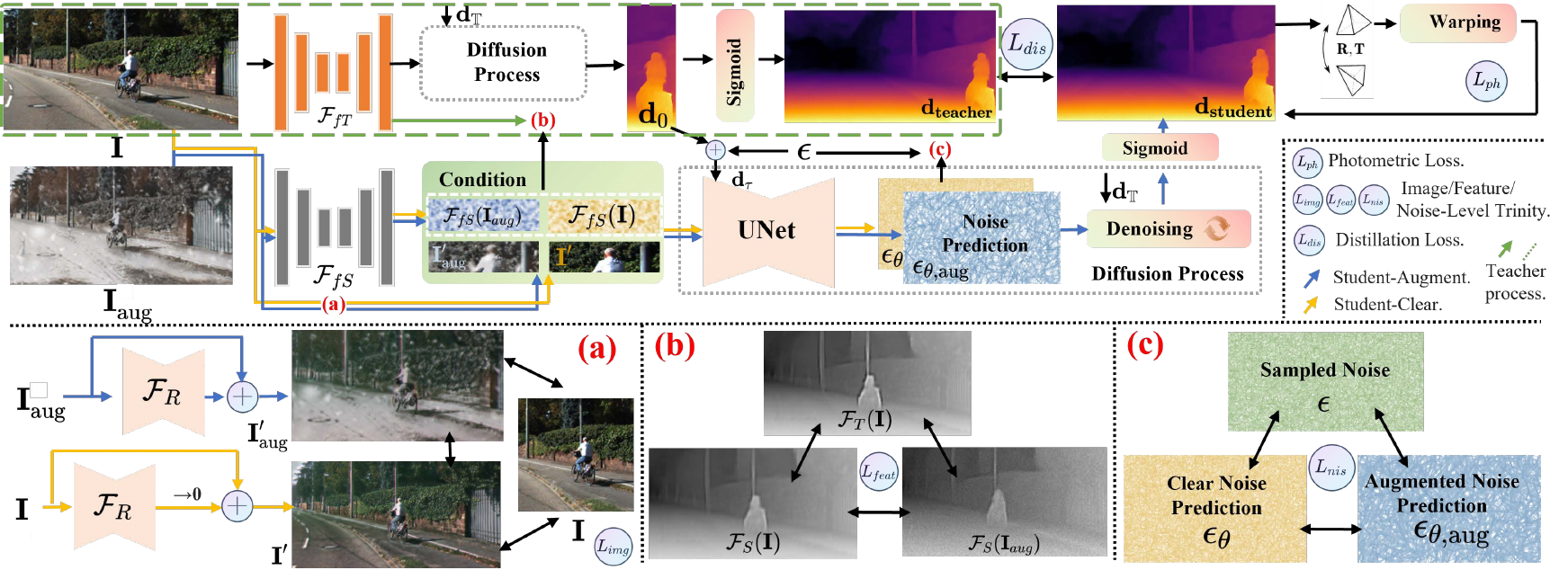}
\vspace{-0.5cm}
\caption{\label{pipline} The training framework of D4RD is depicted in the upper part of the image, where the teacher model $\mathcal{F}_{T}$ is enclosed in the green dashed box, while the student network $\mathcal{F}_{S}$ comprises the remaining parts. There are only \underline{3} components in the whole network: the base feature network (having two symbols, \(\mathcal{F}_{fT}\) in $\mathcal{F}_T$,  \(\mathcal{F}_{fS}\) in $\mathcal{F}_S$), the diffusion process(grey dashed box), and the robust CNN \(\mathcal{F}_R\).
Below that, multi-level trinity learning (\textit{i.e.}, images, depth features, and noise prediction) is presented through (a), (b), and (c), respectively. }
\vspace{-0.4cm}
\end{figure*}

In this section, we will introduce the foundational knowledge of the self-supervised MDE and diffusion-based MDE (Sec. \ref{pre}), the improvements to enhance the stability of the baseline (Sec. \ref{3enhance}), the `trinity' contrast paradigm (Sec. \ref{tlearn}), the multi-level contrasting scheme (Sec. \ref{multi-level}), and the two-stage training strategy (Sec. \ref{overall}).  An overview of the whole framework is shown in Fig. \ref{pipline}. 

\subsection{Preliminaries}
\label{pre}
 To clarify the method, we use abundant symbols. A table of symbol meanings is provided in the supplementary materials for reference.

\textbf{Self-Supervised MDE} uses an auxiliary image $\textbf{I}_{a}$ from the neighboring frame to constrain the output depth with a view-synthesis training pattern. Denoting the MDE model as $\mathcal{F}: \textbf{I} \rightarrow \textbf{d} {\in}  \mathbb{R}^{W \times H }$, we can reach the warped target frame $\textbf{I}_{a \rightarrow t}$ with:
\begin{equation}
\setlength{\abovedisplayskip}{0pt}
\setlength{\belowdisplayskip}{0pt}
\label{proj1}
    \textbf{I}_{a \rightarrow t}=\textbf{I}_{a}\left\langle\operatorname{proj}\left(\textbf{d}, \mathcal{T}_{t \rightarrow a},K\right)\right\rangle ,
\end{equation}
where$\mathcal{T}_{t \rightarrow a}$ denotes the relative camera poses obtained from the pose network, and K denotes the camera intrinsic. 

Afterwards, we can compute the photometric reconstruction loss between \textbf{I} and $\textbf{I}_{a \rightarrow t}$ to constrain the depth:
\begin{equation}
\label{ph}
L_{p h}=\varepsilon_1 \frac{1-\operatorname{SSIM}\left(\textbf{I}, \textbf{I}_{a \rightarrow t}\right)}{2}+\varepsilon_2\left\|\textbf{I}-\textbf{I}_{a \rightarrow t}\right\|.
\end{equation}

In our implementation, DR4D takes the semi-augmented synthesis from Robust-Depth\cite{iccv} to replace Eq. \ref{proj1} with:
\begin{equation}
\label{proj2}
    \textbf{I}_{a \rightarrow t}^\prime=\textbf{I}_{a}\left\langle\operatorname{proj}\left(\textbf{d}_{aug}, \mathcal{T}_{t \rightarrow a},K\right)\right\rangle ,
\end{equation}
where $\textbf{d}_{aug}$ is the depth map inferred form the augmented images and $\textbf{I}_{a \rightarrow t}^\prime$ will replace  $\textbf{I}_{a \rightarrow t}$ in Eq. \ref{ph}.

\textbf{Diffusion-based MDE} redefines the depth estimation task as a conditional denoising process. Given the image $ \textbf{I} \ {\in}\  \mathbb{R}^{W \times H \times 3}$ and the corresponding depth distribution $D$, we can add the Gaussian noise $\epsilon$ to get the noisy sample $\textbf{d}_\tau$:
\begin{equation}
\label{forward1}
{\textbf{d}_\tau } = \sqrt {{{\overline \alpha  }_\tau }} {D} + \sqrt {1 - {{\overline \alpha  }_\tau }} \epsilon , \quad \epsilon  \sim \mathcal{N}\left( {0,I} \right), 
\end{equation}
and iteratively reverse it by removing the predict-added noise $\epsilon_\theta$:
\begin{equation}
\label{denoise}
	{\epsilon_\theta }\left( {{\textbf{d}_{\tau  - 1}}\left| {{\textbf{d}_\tau },\textbf{{c}}} \right.} \right): = \mathcal{N}\left( {{{\mu _\theta }}\left( {{\textbf{d}_\tau },\tau ,\textbf{c}} \right),\sigma _\tau ^2\textbf{n}} \right),  \textbf{c}= \mathcal{F}_f (\textbf{I}) .
\end{equation}
Here $\tau$ consists of $\mathbb{T}$ steps, $\bar{\alpha}_\tau := \prod_{s=1}^{\tau}{1 \!-\! \beta_s}$, 
and $\{\beta_1, \ldots, \beta_{\mathbb{T}} \}$ is the variance noise schedule for $\mathbb{T}$ steps \cite{ddpm}. The $\mathcal{N}(\cdot;\cdot)$ stands for gaussian sampling, $\mathcal{F}_f(\cdot)$ is a feature extractor, and $\sigma _\tau ^2\textbf{n}$ is typically set to \textbf{0} to make the estimation process deterministic.

Obviously, if $\epsilon_\theta$  is completely consistent with the $\epsilon$, we can reverse the depth result precisely, so we have the training loss:
\begin{equation}
\label{ddimloss}
L_{ddim} = \mathbb{E}_{\textbf{d}_0, \epsilon \sim \mathcal{N}(0,1),\tau \sim \mathcal{U}(\mathbb{T})} \left\| \epsilon - {\epsilon_\theta }\left( {{\textbf{d}_{\tau  - 1}}\left| {{\textbf{d}_\tau },\textbf{{c}}} \right.} \right) \right\|^2_2,
\end{equation}
where $\mathcal{U}(\cdot)$ represents uniformly sampling. For self-supervised learning, D4RD adopts the pseudo-GT diffusion as used in MonoDiffusion \cite{monodfs}, which introduces a pre-trained model $\mathcal{F}_{T}$ to infer the pseudo-depth $\textbf{d}_T$ and it will replace $D$ in Eq. \ref{forward1}. 


\subsection{Stable Depth with Diffusion Models}
\label{3enhance}
\textbf{Pseudo-depth knowledge distillation.} 
To improve the performance, MonoDiffusion\cite{monodfs} adopted a weighted $L_1$ loss as the distillation constraint on the predicted depth. The weights are calculated based on the photometric difference using a fixed threshold. However, such a fixed threshold may stick the network to a performance bottleneck. To this end, we replace the fixed filter with an adaptive one:
\begin{equation}
    M=\left[\min _{a} p h\left(\textbf{I}, \textbf{I}_{a \rightarrow t}^{T}\right)<\frac{\lambda}{\text { epoch }}\right],
\end{equation}
 where $\lambda$ is a constant set to 1.5, and $\textbf{I}_{a \rightarrow t}^{T}$ is the warped target image using  $\textbf{d}_T$. 
 This dynamic weight initially enables D4RD to converge with the entire depth map and subsequently filters out the less accurate regions to reduce the negative effects brought by inaccurate pseudo-depth labels.
Moreover, we find that BerHu loss\cite{berhu} yields a lower error than the $L_1$ loss. It imposes a greater penalty on pixels with higher errors using $L_2$ loss, while maintaining the advantage of L1 loss at small errors. In short, we conclude with a dynamically weighted BerHu loss as our constraint for knowledge distillation:
\begin{equation}
    L_{\text {dis }}=M \odot L_{\text {berhu }}(\textbf{d}_{S, aug}, \textbf{d}_{T, clr}) ,
 \end{equation}
 where $\textbf{d}_{T, clr}$ and $\textbf{d}_{S, aug}$ denote the predicted result of the clear or augmented images from $\mathcal{F}_T$ and $\mathcal{F}_S$, respectively.
 
\textbf{Outlier depth removal.} 
As mentioned before, since the iterative denoising process may produce unnatural depths (\textit{e.g.}, negative values), previous methods\cite{cvprdfs} intend to implement diffusion in the latent space:
\begin{equation}
{\textbf{d}_\tau } = \sqrt {{{\overline \alpha  }_\tau }} \mathcal{V}(\textbf{d}) + \sqrt {1 - {{\overline \alpha  }_\tau }} \epsilon ,
\textbf{d}=\mathcal{V}^{-1}(\textbf{d}_0),
\end{equation}
where $\mathcal{V}(\cdot)$ is a pre-trained variational-auto-encoder (VAE), $\mathcal{V}^{-1}(\cdot)$ is the reverse decoder, and $\textbf{d}_0$ no longer represents the final prediction. This manner undoubtedly introduces extra calculations. In this paper, we simplify the encoder and decoder as \textbf{sigmoid}$^{-1}$ and \textbf{sigmoid}, respectively:
\begin{equation}
\label{forward2}
\textbf{d}_\tau  = \sqrt {{{\overline \alpha  }_\tau }} \textbf{sigmoid}^{-1}(\textbf{d}_T) + \sqrt {1 - {{\overline \alpha  }_\tau }} \epsilon,
    \textbf{d}=\textbf{sigmoid}(\textbf{d}_0),
\end{equation}
where $\textbf{sigmoid}(\cdot)$ will transform the outputs into a limited range $[0,1]$ and our experiments show that this simple manner can effectively remove the depth outliers. Eventually, D4RD adopts Eq. \ref{forward2} in the denoising manner.

\textbf{Feature-image joint condition.}
Our experiments have revealed that using the concatenation of the input image and its depth feature as the diffusion condition can bring better performance.  It can be elaborated as:
\begin{equation}
\label{concat}
    \textbf{c}=\textbf{I} \oplus \mathcal{F}_f(\textbf{I}),\quad \mathcal{F}_f(\textbf{I}) \in \mathbb{R}^{1 \times H \times W},
\end{equation}
where $\oplus$ represents the channel-dim concatenation. We visualize the feature map in Fig. \ref{mult-vis} to analyze the reasons and reveal that $\mathcal{F}_f(\textbf{I})$ is a coarse `depth map' with some additional incorrect edges. In other words, $\mathcal{F}_f$ focuses solely on depth-relevant information, and the combination in Eq. \ref{concat} provides higher-level contextual and semantic information, which helps the Fdiffusion architecture in understanding the scenes, predicting noises, and ultimately generating better estimations.


\subsection{Trinity Contrastive learning}
\label{tlearn}

\begin{figure}[ht]
\centering
\includegraphics[width=\linewidth]{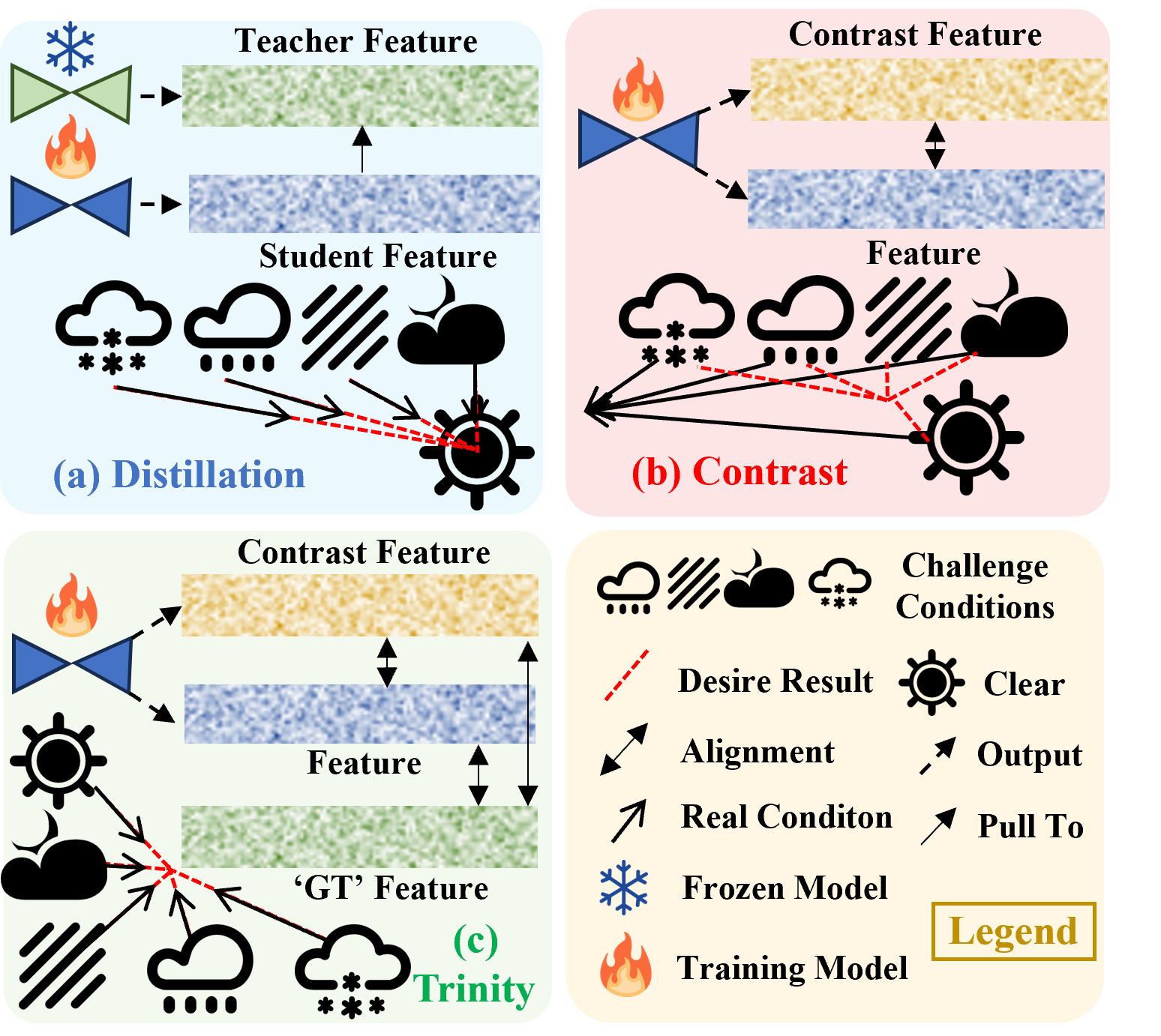}
\vspace{-0.6cm}
\caption{\label{trinity} Visual comparisons among three types of robust learning methods. Our trinity contrast method is more close to our expected effect and achieves better actual performance.}
\vspace{-0.6cm}
\end{figure}

The proposed trinity contrastive scheme integrates both knowledge distillation and contrastive learning.
For the first one, the general pattern is:
\begin{equation}
    L_{dis}=|\mathcal{F}_S(\textbf{I}_{aug})-\underline{\mathcal{F}_T(\textbf{I})}|,
\end{equation}
where $\underline{\text{underline}}$ denotes the prediction from a frozen model. However, Fig. \ref{trinity}(a) has shown that this strategy struggles to teach $\mathcal{F}_S$ when the augmented image $\textbf{I}_{aug}$ has a significant domain difference. Moreover, there is an evident performance upper bound for $\mathcal{F}_S$.
\newline
For the latter, the usual form is:
\begin{equation}
    L_{cst}=| \mathcal{F}(\textbf{I})-\mathcal{F}(\textbf{I}_{aug}) | .
\end{equation}
As shown in Fig. \ref{trinity} (b), without further guidance, simply gathering the depth map can easily lead to network collapse \cite{simsiam}. The self-supervised methods \cite{zcq, iccv} solve this by providing additional photometric guidance, but this guidance is not accurate enough and requires twice the training time (Eq.\ref{ph} is calculated twice for $\textbf{I}_{a \rightarrow t}$ and $\textbf{I}_{a \rightarrow t}^\prime$ separately).

In diffusion models, we discovered that in Eq. \ref{forward1}, the added noise $\epsilon$, which is sampled from $ \mathcal{N}\left( {0, I} \right)$, can serve as a perfect guiding anchor for the estimated noise $\epsilon_\theta$ in Eq. \ref{ddimloss} and do not require any extra annotation. Let $\epsilon_{\theta}$ and $\epsilon_{\theta,\text {aug }}$ represent the noise predictions for the clear image \textbf{I} and augment image $\textbf{I}_{aug}$, respectively.
Once these noises are identical and accurate, the final denoising depth will demonstrate consistency and robustness. Hence, as shown in Fig. \ref{pipline}(c), we reformulate $L_{ddim}$ as a novel noise-level trinity loss:
\begin{equation}
    L_{\text {nis}}=\eta_{1}\|\epsilon_{\theta,\text {aug }}-\epsilon_{\theta}\|_{2}^2+ \\
    \eta_{2}(\|\epsilon_{\theta,\text {aug }}-\epsilon \|_{2}^2+\|\epsilon_{\theta}-\epsilon \|_{2}^2),
\end{equation}
where $\eta_{1}$ and $\eta_{2}$ are set to 0.5 and 1, respectively . With the natural and perfect sampled noise label, the proposed trinity can stably gather the pair of estimated noises toward an optimal anchor position. We also experimentally prove its superiority to the other two traditional strategies in Section \ref{multi-xr}.

\subsection{Multi-level Contrast}
\label{multi-level}

\begin{figure}[ht]
\centering
\hspace*{-1.25em}
\includegraphics[width=1.1\linewidth]{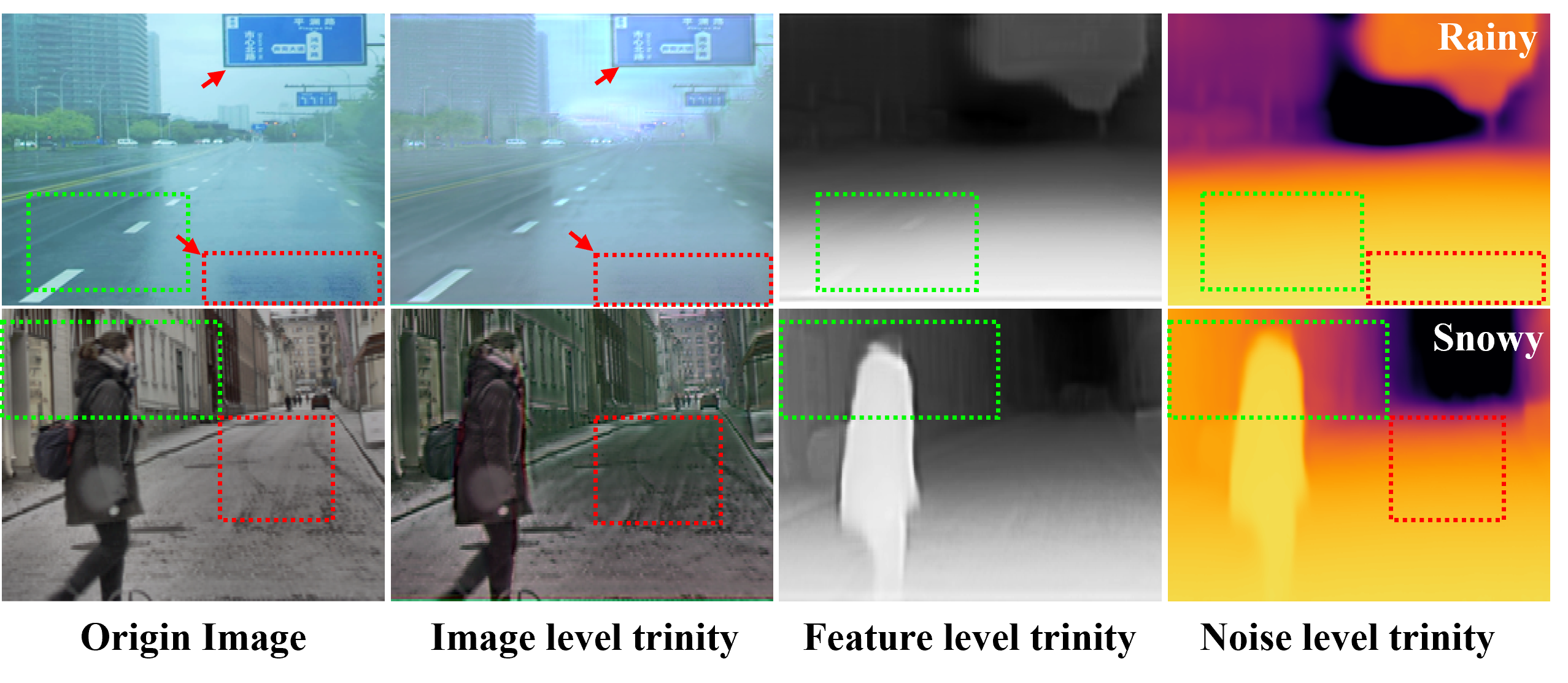}
\vspace{-0.8cm}
\caption{\label{mult-vis}Visual results of each level trinity. Compared to the origin image, as shown in the red dashed rectangle, the image level trinity can assist in handling water surface artifacts and ground snow. The feature level trinity builds a coarse depth map with some wrong edges but the noise trinity fixes them all. Better viewed when zooming in.
}
\vspace{-0.3cm}
\end{figure}

Prior studies\cite{forget} have revealed that the burden of handling complex conditions mainly concentrates on a specific part of the network. Such an unbalanced burden may hinder the potential of robust algorithms. To this end, as shown in Fig. \ref{pipline}, we further extend the noise-level `trinity' approach to more generic targets (\textit{e.g.}, feature and image levels), through which we distribute the responsibility of robust perception across the whole network. 

\textbf{Feature level.} We first extend it to the depth feature level, which exploits the robust ability of feature network $\mathcal{F}_f$. However, at this time, we lack the guidance as perfect as the added noise. Instead, as depicted in Fig. \ref{pipline} (b), we adopt a teacher model $\mathcal{F}_{fT}$ to get the suboptimal guiding label and build the feature-level `trinity' loss as:
\begin{equation}
\begin{array}{l}
L_{feat}=\omega_{1}\|\mathcal{F}_{fS}(\textbf{I})-\mathcal{F}_{fS}(\textbf{I}_{aug})\|_{1}+ \\
\omega_{2}(\|\mathcal{F}_{fT}(\textbf{I})-\mathcal{F}_{fS} (\textbf{I})\|_{1}+\|\mathcal{F}_{fT}(\textbf{I})-\mathcal{F}_{fS}(\textbf{I}_{aug})\|_{1}),
\end{array}
\end{equation}
where $\omega_1$ and $\omega_2$ are empirically set to 1 and 0.5, respectively. 

\textbf{Image level:}
In addition to noise and feature, we also implement an image-level trinity contrast by setting the guiding label as the clear image \textbf{I}. As depicted in Fig. \ref{pipline}(a), we design a simple CNN network that performs down-up sampling (denoted as $\mathcal{F}_R$) to get the enhanced image via:
\begin{equation}
    \left\{\begin{array}{l}
\textbf{I}_{\text {aug }}^{\prime}=\mathcal{F}_{R}\left(\textbf{I}_{\text {aug }}\right)+\textbf{I}_{\text {aug }} \\
\textbf{I}^{\prime}=\mathcal{F}_{R}(\textbf{I})+\textbf{I}
\end{array}\right..
\end{equation}
Then the image-level trinity is formulated as:
\begin{equation}
L_{img}=\delta_{1}(\|\textbf{I}_{\text {aug }}^{\prime}-\textbf{I}\|_{1}+\|\textbf{I}^{\prime}-\textbf{I}\|_{1})+\delta_{2}\|\textbf{I}_{\text {aug }}^{\prime}-\textbf{I}^{\prime}\|_{1},
\end{equation}
where the values of $\delta_{1}$ and $\delta_{2}$ are same as those of $\omega_1$ and $\omega_2$.

As shown in Fig. \ref{mult-vis}, through three trinity contrasts from different levels, the stresses of robust perception are evenly distributed among three parts of the network. They encourage the model to recognize and resist various types of scene degeneration at multiple stages, thus boosting the overall prediction reliability.

\begin{table*}[ht]
\caption{\label{kittires}\textbf{Quantitative results on WeatherKITTI and KITTI dataset.} 
For the \colorbox{low}{error-based metrics}, the lower value is better; and for the \colorbox{high}{accuracy-based metrics}, the higher value is better. The best and second-best results are marked in \textbf{bold} and \underline{underline}.}\label{table:results} 
\vspace{-0.3cm}
\resizebox{\linewidth}{!}{
\begin{tabular}{c||ccccccc|ccccccc}
\hline
                         & \multicolumn{7}{c|}{WeatherKITTI}& \multicolumn{7}{c}{KITTI}\\ 
                         \cline{2-15} 
\multirow{-2}{*}{Method} & \cellcolor{low}AbsRel & \cellcolor{low}SqRel & \cellcolor{low}RMSE & \multicolumn{1}{c|}{\cellcolor{low}RMSElog} & \cellcolor{high}$ a_1 $ & \cellcolor{high}$ a_2 $ & \cellcolor{high}$ a_3 $ & \cellcolor{low}AbsRel & \cellcolor{low}SqRel & \cellcolor{low}RMSE & \multicolumn{1}{c|}{\cellcolor{low}RMSElog} & \cellcolor{high}$ a_1 $ & \cellcolor{high}$a_2 $ & \cellcolor{high}$ a_3 $ \\ \hline
MonoDiffusion\cite{monodfs}&0.133& 0.960& 5.212& \multicolumn{1}{c|}{0.213}& 0.828& 0.945& \multicolumn{1}{c|}{0.977} &0.103& 0.718& 4.411& \multicolumn{1}{c|}{0.178}& 0.894& 0.965& \underline{0.984}\\
MonoViT\cite{vit} & 0.120 & 0.899 & 5.111 & \multicolumn{1}{c|}{0.200} & 0.857 & 0.953 & \multicolumn{1}{c|}{0.980} & 0.099 & 0.708 & 4.372 & \multicolumn{1}{c|}{0.175} & \textbf{0.900} & \underline{0.967} & \underline{0.984} \\
Robust-Depth\cite{iccv} & 0.107 & 0.791 & 4.604 & \multicolumn{1}{c|}{0.183} & 0.883 & 0.963 & \multicolumn{1}{c|}{\underline{0.983}} & 0.100 & 0.747 & 4.455 & \multicolumn{1}{c|}{0.177} & 0.895 & 0.966 & \underline{0.984} \\
EC-Depth\cite{ecdp} & 0.110 & 0.790 & 4.744 & \multicolumn{1}{c|}{0.185} & 0.875 & 0.960 & \multicolumn{1}{c|}{\underline{0.983}} & 0.100 & \underline{0.689} & \underline{4.315} & \multicolumn{1}{c|}{\underline{0.173}} & 0.896 & \underline{0.967} & \textbf{0.985} \\
Robust-Depth*\cite{iccv} & 0.103 & 0.806 & 4.517 & \multicolumn{1}{c|}{0.180} & 0.894 & 0.964 & \multicolumn{1}{c|}{\underline{0.983}} & 0.100 & 0.776 & 4.440 & \multicolumn{1}{c|}{0.177} & \underline{0.900} & 0.965 & 0.983 \\
WeatherDepth\cite{wjy} & 0.103 & \underline{0.738} & 4.414 & \multicolumn{1}{c|}{0.178} & 0.892 & 0.965 & \multicolumn{1}{c|}{\textbf{0.984}} & 0.099 & 0.698 & 4.330 & \multicolumn{1}{c|}{0.174} & 0.897 & \underline{0.967} & \underline{0.984} \\
EC-Depth*\cite{ecdp} & \underline{0.102} & 0.762 & \underline{4.400} & \multicolumn{1}{c|}{0.177} & \underline{0.895} & \textbf{0.967} & \multicolumn{1}{c|}{\textbf{0.984}} & \underline{0.098} & 0.732 & 4.326 & \multicolumn{1}{c|}{0.174} & \textbf{0.902} & \textbf{0.968} & \underline{0.984} \\ \hline
D4RD & \textbf{0.099} & \textbf{0.688} & \textbf{4.377} & \multicolumn{1}{c|}{\textbf{0.174}} & \textbf{0.897} & \underline{0.966} & \multicolumn{1}{c|}{\textbf{0.984}} & \textbf{0.097} & \textbf{0.665} & \textbf{4.312} & \multicolumn{1}{c|}{\textbf{0.171}} & \underline{0.900} & \underline{0.967} & \textbf{0.985} \\ \hline

\end{tabular}
}
\vspace{-0.4cm}
\end{table*}

\begin{figure*}[ht]
\centering
\includegraphics[width=0.95\linewidth]{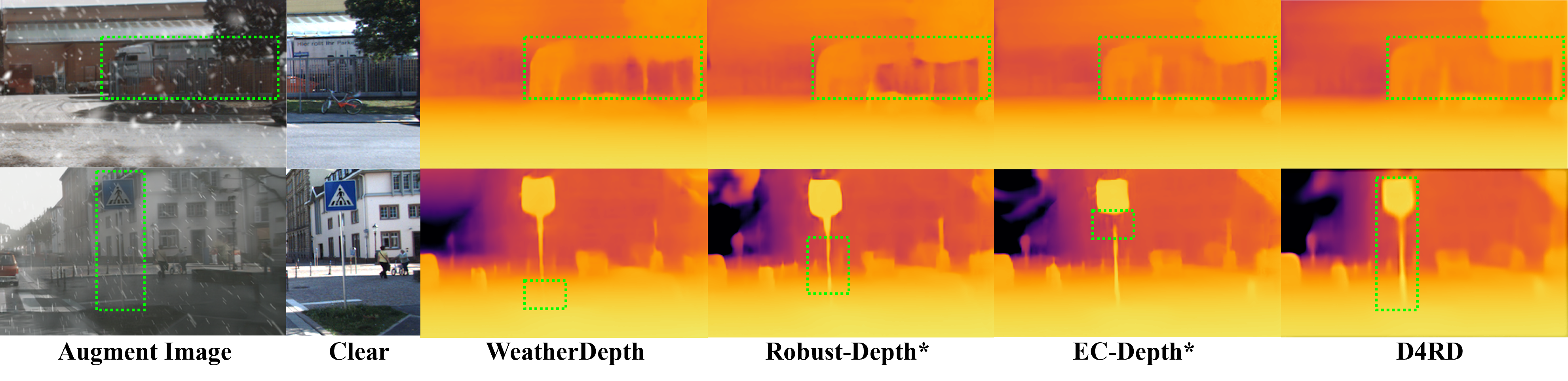}
\vspace{-0.6cm}
\caption{\label{kittivis}Qualitative results for the WeatherKITTI dataset. We compare D4RD with the current SoTA RMDE methods in the adverse rain and snow subsets. The part marked with `Clear' is the corresponding sunny image (processed for more clarity). Regions with prominent differences are highlighted using dashed boxes. }
\vspace{-0.4cm}
\end{figure*}

\subsection{Two stage training}
\label{overall}

\textbf{Motivations.} 
We implement a two-step training strategy for two reasons.
First, the proposed multi-level trinity contrastive paradigm requires the condition labels from the teacher model, but it does not exist initially. On the other hand, the teacher model $\mathcal{F}_T$ from the second stage can integrate the historical knowledge of stage one, yielding more accuracy and reliable pseudo-depth. 

\textbf{The first training phase.}
As shown in Fig. \ref{pipline}, We feed the clear image $\textbf{I}$ and its augmentation $\textbf{I}_{aug}$ as a mini-batch to the model and obtain their depth feature maps. Then we concat them with their corresponding images with Eq. \ref{concat} to get the $\textbf{c}_{aug}$ and  \textbf{c}. Next, both of them will be delivered to diffusion UNet to predict $\epsilon_{\theta}$/$\epsilon_{\theta,\text {aug }}$ and calculate the noise-level trinity loss $L_{nis}$. Note only $\epsilon_{\theta,\text {aug}}$ will be used for the denoising process and $L_{dis}$, $L_{ph}$ will be imposed on \textbf{d}, which implies the proposed trinity will not double the computational burden as the traditional contrastive learning. With an extra edge-aware constraint $L_e$ from \cite{mdp2}, we finally conclude the total loss at the first stage:
\begin{equation}
    L_{stage1}= L_{nis}+L_{dis}+L_{ph}+\rho L_{e},
\end{equation}
where $\rho$ is empirically set to 1e-3.

\textbf{The second training phase.}
The well-trained model from the first stage, now equipped with fixed weights, is designated as $\mathcal{F}_T$ for the second stage. We train $\mathcal{F}_S$ from scratch and further add $L_{feat}$ and $L_{img}$ to build the multi-level contrasts as mentioned in Section \ref{multi-level}. The total loss of the second stage can be written as:
\begin{equation}
    L_{stage2}= L_{nis}+L_{dis}+L_{ph}+\rho L_{e}+\eta(L_{feat} + L_{img}),
\end{equation}
where $\eta$ is empirically set to 0.5.


\begin{figure*}[ht]
\centering
\includegraphics[width=0.95\linewidth]{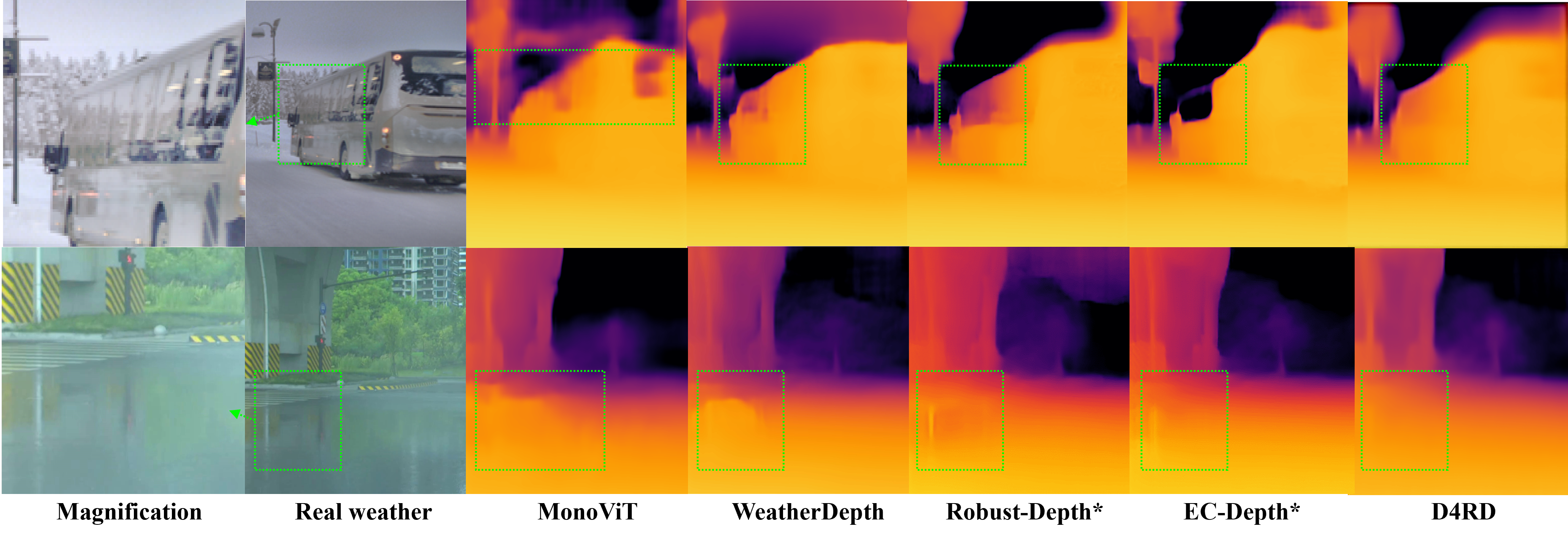}
\vspace{-0.5cm}
\caption{\label{realvis}Qualitative results for the real weather dataset. Besides the RMDE methods, we also compare D4RD with the standard method MonoViT\cite{vit} in snowy and rainy environments to show our method's practicality. The part marked with 'Magnification' is processed for more clarity.}
\vspace{-0.4cm}
\end{figure*}

\section{Experiment}
\subsection{Dataset}
\label{dataset}
\textbf{WeatherKITTI}\cite{wjy} is built on the KITTI dataset and includes 6 kinds of weather augmentation. We train D4RD on this dataset because, compared with other synthetic datasets\cite{iccv, eccv}, it is more realistic in simulating real-world complex scenarios. WeatherKITTI shares the same train/validate/test splits as KITTI. Therefore, for training, we follow Zhou's split\cite{zhou} (containing 39,810 training images and 44,24 validation images), and for testing, we adopt Eigen's split\cite{eigen} with 697 test images.

\textbf{KITTI-C}\cite{Kong2023RoboDepthRO} is a synthetic dataset built for the RoboDepth Competition. In total, the benchmark contains 18 types of perturbations with 5 levels of corruption magnitudes, most of which are stronger than WeatherKITTI. We use it to fairly compare D4RD with their competition champions\cite{ecdp} to further demonstrate our robust performance.

\textbf{DrivingStereo}\cite{stereo} is a large-scale real-world dataset. It contains four subsets of images under four weather conditions (\textit{i.e.}, fog, cloudy, rainy, and sunny), with each subset containing 500 images. We test RMDE models on this dataset to characterize their ability to handle complex real-world conditions.

\textbf{Dense}\cite{dense} is one of the few real snowy datasets and it has approximately 1,500 snowy images in total. We introduce this benchmark to test snowy scenes. To ensure consistency with the above datasets, we followed \cite{wjy} and sampled 1 image every 3 to obtain 500 images as test data. To generate accurate depth GT in challenging snowy conditions, we use the DROR\cite{dror} algorithm to remove inaccurate depth readings from LiDARs affected by snowflakes. In addition, a small portion of distorted areas has been trimmed, resulting in a final resolution of $1880 \times 924$. We will explain the reasons for not using the snowy part of the CADC dataset in the supplementary materials.

\subsection{Implementation Details}
\label{implement}
In this paper, D4RD is based on MonoDiffusion\cite {monodfs}, which is currently the only diffusion-based self-supervised MDE pipeline that combines the diffusion architecture with Lite-Mono\cite{litemono}. To maintain consistency with previous RMDE researches\cite{iccv,wjy,ecdp}, we replace Lite-Mono with MonoViT\cite{vit} in our D4RD. 

D4RD is implemented in PyTorch and trained on a single NVIDIA RTX 3090. We use the AdamW optimizer\cite{adamw} and other similar hyperparameters as MonoViT. The batch size is set to 6, and the training process runs 30 epochs in each stage. Considering the resolution differences across the datasets, we resize them to $640 \times 192$ for both training and testing. For the diffusion process, we follow MonoDiffusion\cite{monodfs} and set it with 1000 training steps and 20 inference steps.

For fair comparisons, we retrained Robust-Depth\cite{iccv} and EC-Depth\cite{ecdp} on the WeatherKITTI dataset. Furthermore, to demonstrate D4RD's ability to handle robust scenes, we also retrain our model with the same training data used by \cite{ecdp}, named D4RD$^{\dagger}$, to compare the performance on the KITTI-C dataset.

\vspace{-0.15cm}
\subsection{Quantitative and Qualitative Comparison}
Here, we adopt the same evaluation metrics and errors up to 80m for all comparisons as in \cite{mdp2}. Please refer to the supplementary materials for more details. The comparative methods include the baseline MonoDiffusion, traditional RMDE models' baseline MonoViT\cite{vit}, previous SoTA RMDE models WeatherDepth\cite{iccv}, EC-Depth\cite{ecdp}, Robust-Depth\cite{iccv}, and their retrained models. Besides, \cite{iccv} and \cite{wjy} apply their approaches to various baselines, we want to clarify that we choose the one that uses MonoViT\cite{vit}. In \cite{ecdp}, they take a two-stage training pipeline and we select the second stage model for the comparison.

\textbf{WeatherKITTI result.}
We show detailed comparative experiments on the KITTI and WeatherKITTI datasets in Table \ref{kittires}, where our method outperforms other SoTA methods in all metrics by a large margin. Specifically, D4RD demonstrates a 25.6\% and 17.5\% decrease in AbsRel errors compared to the baselines(MonoDiffusion and MonoViT) under challenging conditions (WeatherKITTI). Furthermore, its performance either exceeds or closely matches that of these baselines in standard conditions (KITTI). The qualitative results are shown in Fig. \ref{kittivis}. From the regions highlighted in green boxes, the compared schemes are affected by weather particle occlusion and blurs. In contrast, D4RD can robustly mine the consistent depth plane of the truck body and accurately estimate the correct contour of the road sign.

\begin{table}[!tbp]
\setlength{\abovecaptionskip}{0cm}
\setlength{\belowcaptionskip}{-0.4cm}
\caption{\label{zeroshot}\textbf{Zero-shot evaluation on the Dense and DrivingStereo dataset.}}
\hspace*{-1.5em}
\resizebox{1.1\linewidth}{!}{
\begin{tabular}{c||cccc|ccc}
\hline
Method    & \cellcolor{low}AbsRel & \cellcolor{low}SqRel & \cellcolor{low}RMSE & \multicolumn{1}{c|}{\cellcolor{low}RMSElog} & \cellcolor{high}$ a_1 $ & \cellcolor{high}$ a_2 $ & \cellcolor{high}$ a_3 $   \\ 
\hline
\multicolumn{8}{c}{(a)DrivingStereo Sunny}\\
\hline
MonoDiffusion\cite{monodfs}&     0.162&     1.738&     7.535&     0.220&     0.799&     0.936&     0.975\\
MonoViT\cite{vit} & \underline{0.150} & 1.615 & 7.657 & 0.211 & \textbf{0.815} & 0.943 & \underline{0.979} \\
Robust-Depth\cite{iccv} & 0.152 & 1.574 & 7.293 & 0.210 & 0.812 & \underline{0.944} & \underline{0.979} \\
EC-Depth\cite{ecdp} & 0.151 & \textbf{1.436} & \underline{7.213} & \underline{0.209} & 0.808 & \underline{0.944} & \underline{0.979} \\
Robust-Depth*\cite{iccv} & 0.152 & 1.651 & 7.369 & 0.212 & 0.810 & \underline{0.944} & 0.978 \\
WeatherDepth\cite{wjy} & 0.155 & 1.562 & 7.356 & 0.213 & 0.803 & 0.941 & 0.978 \\
EC-Depth*\cite{ecdp} & 0.155 & 1.538 & 7.301 & 0.211 & 0.804 & 0.941 & \textbf{0.980} \\
\hline
D4RD & \textbf{0.149} & \underline{1.437} & \textbf{7.121} & \textbf{0.207} & \textbf{0.815} & \textbf{0.946} & \textbf{0.980} \\
\hline
\multicolumn{8}{c}{(b)DrivingStereo Cloudy}\\
\hline  
MonoDiffusion\cite{monodfs}&     0.155&     1.853&     7.822&     0.214&     0.807&     0.939&     0.978\\
MonoViT\cite{vit} & \textbf{0.141} & 1.626 & 7.550 & 0.201 & \textbf{0.831} & \underline{0.948} & 0.981 \\
Robust-Depth\cite{iccv} & 0.148 & 1.781 & 7.472 & 0.204 & 0.825 & 0.947 & 0.981 \\
EC-Depth\cite{ecdp} & 0.147 & 1.561 & \underline{7.301} & \underline{0.201} & 0.825 & 0.947 & \textbf{0.983} \\
Robust-Depth*\cite{iccv} & 0.147 & 1.749 & 7.486 & 0.205 & 0.823 & 0.946 & \underline{0.982} \\
WeatherDepth\cite{wjy} & \underline{0.144} & \textbf{1.549} & \underline{7.349} & 0.201 & 0.822 & \textbf{0.949} & \textbf{0.983} \\
EC-Depth*\cite{ecdp} & 0.150 & 1.655 & 7.517 & 0.204 & 0.819 & 0.945 & \underline{0.982} \\
\hline
D4RD & \textbf{0.141} & \underline{1.560} & \textbf{7.271} & \textbf{0.198} & \underline{0.830} & \underline{0.948} & \textbf{0.983} \\
\hline
\multicolumn{8}{c}{(c)DrivingStereo Rainy}\\
\hline  
MonoDiffusion\cite{monodfs}&     0.196&     2.629&    10.546&     0.254&     0.691&     0.905&     0.973\\
MonoViT\cite{vit} & 0.175 & 2.136 & 9.618 & 0.232 & 0.730 & 0.931 & 0.979 \\
Robust-Depth\cite{iccv} & 0.167 & 2.019 & 9.157 & 0.221 & 0.755 & 0.938 & 0.982 \\
EC-Depth\cite{ecdp} & \underline{0.162} & \underline{1.746} & \textbf{8.538} & 0.212 & 0.755 & \textbf{0.947} & \textbf{0.986} \\
Robust-Depth*\cite{iccv} & 0.173 & 2.154 & 9.452 & 0.226 & 0.733 & 0.934 & 0.982 \\
WeatherDepth\cite{wjy} & \textbf{0.158} & 1.833 & 8.837 & \underline{0.211} & \underline{0.764} & 0.945 & \underline{0.985} \\
EC-Depth*\cite{ecdp} & \underline{0.162} & 1.810 & 8.792 & 0.215 & 0.744 & 0.943 & \underline{0.985} \\
\hline
D4RD & \textbf{0.158} & \textbf{1.722} & \underline{8.584} & \textbf{0.208} & \textbf{0.773} & \underline{0.946} & \underline{0.985} \\
\hline
\multicolumn{8}{c}{(d)DrivingStereo Foggy}\\
\hline  
MonoDiffusion\cite{monodfs}&     0.128&     1.540&     8.687&     0.191&     0.831&     0.955&     0.986\\
MonoViT\cite{vit} & \underline{0.109} & 1.204 & 7.760 & 0.167 & 0.870 & 0.967 & 0.990 \\
Robust-Depth\cite{iccv} & \textbf{0.105} & \underline{1.132} & 7.273 & 0.158 & \underline{0.882} & \underline{0.974} & 0.992 \\
EC-Depth\cite{ecdp} & \textbf{0.105} & \textbf{1.061} & \underline{7.121} & \underline{0.155} & 0.880 & \underline{0.974} & \textbf{0.994} \\
Robust-Depth*\cite{iccv} & 0.111 & 1.240 & 7.536 & 0.163 & 0.873 & 0.971 & 0.992 \\
WeatherDepth\cite{wjy} & 0.110 & 1.195 & 7.323 & 0.160 & 0.878 & 0.973 & 0.992 \\
EC-Depth* \cite{ecdp}& 0.111 & 1.177 & 7.315 & 0.160 & 0.870 & \underline{0.974} & \underline{0.993} \\
\hline
D4RD & \textbf{0.105} & \textbf{1.061} & \textbf{7.102} & \textbf{0.154} & \textbf{0.883} & \textbf{0.975} & \textbf{0.994} \\ 
\hline
\multicolumn{8}{c}{(e)Dense Snowy}\\
\hline  
MonoDiffusion\cite{monodfs}&     0.173&     2.169&    10.029&     0.273&     0.744&     0.898&     0.955\\
MonoViT\cite{vit} & 0.162 & 2.063 & 9.787 & 0.262 & 0.762 & 0.904 & 0.960 \\
Robust-Depth\cite{iccv} & 0.157 & 1.992 & 8.945 & 0.240 & \underline{0.786} & \underline{0.923} & \underline{0.971} \\
EC-Depth\cite{ecdp} & \underline{0.155} & \textbf{1.866} & 8.828 & 0.237 & 0.780 & 0.922 & \textbf{0.972} \\
Robust-Depth*\cite{iccv} & 0.157 & 2.050 & 8.951 & 0.243 & 0.785 & 0.921 & 0.969 \\
WeatherDepth\cite{wjy} & 0.157 & 2.000 & 9.021 & 0.243 & 0.781 & 0.919 & 0.968 \\
EC-Depth*\cite{ecdp} &\textbf{0.154}&1.984&\underline{8.806}&\textbf{0.235}&0.782&\underline{0.923}&\textbf{0.972}\\
\hline
D4RD & \textbf{0.154} & \textbf{1.868} & \textbf{8.780} & \underline{0.236}& \textbf{0.789} & \textbf{0.925} & \textbf{0.972} \\ 
\hline  
\multicolumn{8}{c}{(f)Average}\\
\hline 
MonoDiffusion\cite{monodfs}&0.163&1.986&8.924&0.230&0.774&0.927&0.973\\
MonoViT\cite{vit}        & 0.147    & 1.729  & 8.474  & 0.215    & 0.802         & 0.939            & 0.978     \\
Robust-Depth\cite{iccv}    & 0.146    & 1.700  & 8.028  & 0.207    & \underline{0.812}         & 0.945            & 0.981    \\
EC-Depth\cite{ecdp}       & \underline{0.144}    & \underline{1.534}  &\underline{7.800}  & \underline{0.203}    & 0.810         & \underline{0.947}            & \textbf{0.983 }   \\
Robust-Depth*\cite{iccv}   & 0.148    & 1.769  & 8.159  & 0.210    & 0.805         & 0.943            & 0.981     \\
WeatherDepth\cite{wjy}    & 0.145    & 1.628  & 7.977  & 0.206    & 0.810         & 0.945            & 0.981      \\
EC-Depth*\cite{ecdp}     & 0.146    & 1.633  & 7.946  & 0.205    & 0.804         & 0.945            &\textbf{0.983 }     \\
\hline
D4RD         & \textbf{0.141}    & \textbf{1.530}  & \textbf{7.772 } & \textbf{0.201}    & \textbf{0.818}         & \textbf{0.948}            & \textbf{0.983}            \\ \hline
\end{tabular}
}
\vspace{-4pt}
\end{table}

\begin{table}[!htbp]
\setlength{\abovecaptionskip}{0.1cm}
\caption{\label{cres}\textbf{Evaluation on the KITTI-C dataset.}}
\label{table:drivingstereo}
\hspace*{-1.5em}
\resizebox{1.1\linewidth}{!}{
\begin{tabular}{c||cccc|ccc}
\hline
Method    & \cellcolor{low}AbsRel & \cellcolor{low}SqRel & \cellcolor{low}RMSE & \multicolumn{1}{c|}{\cellcolor{low}RMSElog} & \cellcolor{high}$ a_1 $ & \cellcolor{high}$ a_2 $ & \cellcolor{high}$ a_3 $   \\ 
\hline  
MonoViT\cite{vit} & 0.161 & 1.292 & 6.029 & 0.247 & 0.768 & 0.915 & 0.964 \\
Robust-Depth\cite{iccv} & 0.123 & 0.957 & 5.093 & 0.202 & 0.851 & 0.951 & 0.979 \\
EC-Depth\cite{ecdp} & \underline{0.111} & \underline{0.807} & \textbf{4.561} & \underline{0.185} & \underline{0.874} & \underline{0.960} & \underline{0.983} \\
D4RD$^{\dagger}$ & \textbf{0.108} & \textbf{0.778} & \underline{4.652} & \textbf{0.183} & \textbf{0.880} & \textbf{0.961} & \textbf{0.983} \\
\hline
\end{tabular}
}
\vspace{-16pt}
\end{table}

\begin{table}[!hb]
\setlength{\abovecaptionskip}{0.1cm}
\setlength{\belowcaptionskip}{-0.1cm}
\centering
\caption{\label{compute}Comparison of computation and time usage.}
\hspace*{-1em}
\scalebox{0.8}{\begin{tabular}{c|| c c c c c c}
    \hline
     Model& \cellcolor{low}FLOPs(G)&\cellcolor{low}Param(M)&\cellcolor{low}Memory(GB)&\cellcolor{low}Train/Infer(s/batch)& \cellcolor{low}AbsRel\\
    \hline
    Monovit[\textcolor[rgb]{0,1,0}{36}] & 715.4 & 27.8 & 15.3 & 0.64/0.14 & 0.120\\
    D4RD &881.9&27.9&23.5&1.02/0.32&0.099\\
    \hline
\end{tabular}}
\label{tab1}
\vspace{-15pt}
\end{table}

\textbf{Real Weather Scenes Results.}
To validate the practicality of D4RD, we conduct zero-shot evaluations on real-world challenging datasets. As shown in Table \ref{zeroshot} (a), D4RD still exhibits the best performance on the out-of-distribution clear dataset, demonstrating that our model has a stronger zero-shot ability than previous RMDE methods. Subsequently, we test the robustness of our model using the most common real-world changing conditions, including cloudy, rainy, foggy, and snowy scenes. As Table \ref{zeroshot} shows, D4RD consistently demonstrates the SoTA performance in every scenario (b-e), and (f) further reports that our average performance has improved by a level compared to other methods in AbsRel( 1.41 to around 1.45), which strongly proves the superiority and practicality of our method. Fig. \ref{realvis} displays the qualitative results of our method on real rainy and snowy environments. It can be observed that our method distinguishes the snow scene from the white bus (its windows even reflect some snow scenes) and better identifies artifacts caused by water surface reflections.

\textbf{KITT-C Result.}
In addition to dealing with weather conditions, another role of RMDE is to resist image corruption and perturbations. The first track of RoboDepth competitions\cite{Kong2023RoboDepthRO} focused on solving this problem. The solution from EC-Depth\cite{ecdp} won 1st place among dozens of participants and Table \ref{cres} shows its giant improvement from MonoViT. However, D4RD$^{\dagger}$ further defeats the EC-Depth solution by simply replacing the training dataset with EC-Depth's dataset, which powerfully proves that D4RD is currently the best RMDE framework. 

\subsection{Ablation Study}
In this section, we conduct detailed ablation studies on WeatherKITTI and real-world benchmarks to demonstrate the effectiveness of the proposed components. In the end, we discuss the computational analysis and training costs of D4RD. 

\begin{table*}[ht]
\caption{\label{xr1}\textbf{Ablation study on stable depth with diffusion.} All experiments are implemented in stage one, and the meaning of the abbreviations is as follows: PDE stands for Pseudo-depth Distillation Enhancement; ODR stands for Outlier Depth Removing; FIC stands for Feature-Image joint Condition.}
\label{ablation}  
\vspace{-0.3cm}
\resizebox{0.85\linewidth}{!}{
\begin{tabular}{c|c|cccc||cccc|ccc}
\hline
Benchmark&ID& PDE & ODR & FIC & Trinity & \cellcolor{low}AbsRel & \cellcolor{low}SqRel & \cellcolor{low}RMSE &\cellcolor{low}RMSElog& \cellcolor{high}$ a_1 $ & \cellcolor{high}$ a_2 $ & \cellcolor{high}$ a_3 $ \\
 \hline
\multirow{5}{*}
{WeatherKITTI}  &1&              &  &  &  & 0.106 & 0.780 & 4.548 & 0.181 & 0.888 & 0.964 & 0.983 \\
                &2& $\checkmark$ &  &  &  & 0.104 & 0.775 & 4.502 & 0.179 & 0.892 & 0.965 & 0.983 \\
                &3& $\checkmark$ & $\checkmark$ &  &  & 0.103 & 0.708 & 4.433 & 0.178 & 0.891 & 0.965 & \textbf{0.984} \\
                &4& $\checkmark$ & $\checkmark$ & $\checkmark$ &  & 0.101 & 0.712 & 4.423 & 0.177 & 0.895 & 0.965 & \textbf{0.984} \\ 
                &5& $\checkmark$ & $\checkmark$ & $\checkmark$ & $\checkmark$ & \textbf{0.100} & \textbf{0.696} & \textbf{4.375} & \textbf{0.175} & \textbf{0.896} & \textbf{0.966} & \textbf{0.984} \\    
\hline
\end{tabular}
}
\vspace{-0.3cm}
\end{table*}

\begin{table*}[ht]
\caption{\label{xr2}\textbf{Ablation study on contrast manner and level.} All experiments are implemented in the second stage and the meaning of the abbreviations is as follows. NIS: the Noise level contrast; DF: the Depth Feature level contrast; IMG: the Image level contrast. Real Weather shows the average test result across all kinds of real weather conditions in Tab.\ref{zeroshot}}
\label{ablation}  
\vspace{-0.3cm}
\resizebox{0.85\linewidth}{!}{
\begin{tabular}{c|c|cccc||ccccccc}
\hline
Benchmark &ID& NIS & DF & IMG & Method & \cellcolor{low}AbsRel & \cellcolor{low}SqRel & \cellcolor{low}RMSE &\cellcolor{low}RMSElog& \cellcolor{high}$ a_1 $ & \cellcolor{high}$ a_2 $ & \cellcolor{high}$ a_3 $ \\
 \hline
 \multirow{6}{*}
{WeatherKITTI}
&1& & & &                                  & 0.100   & 0.696  & 4.375 & 0.175    & 0.896  & 0.966  & 0.984            \\
&2&$\checkmark$ & & &  Trinity             & 0.100   & 0.700  & 4.367 & 0.175    & 0.897  & 0.966  & 0.984            \\
&3&$\checkmark$ & $\checkmark$&  & Trinity & \textbf{0.099}   & 0.698  & \textbf{4.364} & 0.175    & \textbf{0.898}  & 0.966  & 0.984            \\
&4&$\checkmark$& $\checkmark$& $\checkmark$&Distill& 0.100   & 0.704  & 4.384 & 0.175    & 0.896  & 0.966  & 0.984  \\
&5&& $\checkmark$& $\checkmark$&Contrast&0.100&0.712&4.425&0.176&0.895&0.965&0.984\\
&6&$\checkmark$& $\checkmark$& $\checkmark$&Trinity &\textbf{0.099}   & \textbf{0.688}  & 4.377 & \textbf{0.174}    & 0.897  & \textbf{0.966}  & \textbf{0.984} \\
\hline
 \multirow{6}{*}
{Real Weather}
&7& & & &                                             & 0.143   & 1.566  & 7.853 & 0.202    & 0.815  & 0.947 & 0.982  \\ 
&8&$\checkmark$ & & &  Trinity                        & \textbf{0.141}   & 1.540  & 7.851 & \textbf{0.201}    & 0.817  & \textbf{0.948} & \textbf{0.983}  \\
&9&$\checkmark$ & $\checkmark$&  & Trinity            & \textbf{0.141}   & 1.535  & 7.850 & 0.202    & 0.817  & 0.947 & 0.982  \\
&10&$\checkmark$& $\checkmark$& $\checkmark$&  Distill & 0.145   & 1.588  & 7.894 & 0.204    & 0.812  & 0.946 & 0.981  \\
&11&&$\checkmark$ & $\checkmark$&  Contrast& 0.142&1.538&7.835&0.202&0.815&\textbf{0.948}&0.982\\
&12&$\checkmark$& $\checkmark$& $\checkmark$& Trinity  & \textbf{0.141}   & \textbf{1.530}  & \textbf{7.772} & \textbf{0.201}    & \textbf{0.818}  & \textbf{0.948} & \textbf{0.983}  \\
\hline
\end{tabular}
}
\vspace{-0.3cm}
\end{table*}

\textbf{Effectivness of Diffusion Enhancement.}
\label{3enhance-xr}
Table \ref{xr1} shows the effectiveness of each enhancement based on the first stage of training. We take the MonoDiffusion (replaced with MonoViT) as the baseline (ID 1). Firstly, the distillation enhancement (ID 2) is added and produces a significant improvement. This gain is reasonable because, compared with MonoDiffusion which only considers clear scenes, the depth estimation in complex scenes typically requires more powerful distillation. Next, we add outlier depth removal (ID 3), which brings a notable improvement in RMSE and SqRel metrics because the unnatural depth values have a more significant impact on the second-order indicators. Finally, we improve the condition (ID 4). With the introduction of image context information, the diffusion-based model further shows an improvement in the $a_1$ metric representing estimation stability.

\textbf{Effectiveness of Trinity Contrast.}
\label{trinity-xr}
As reported in Table \ref{xr1} (ID 5), the incorporation of trinity in the first stage model produces overall performance gains in all aspects. Meanwhile, in Table \ref{xr2}, we can compare the performance of three paradigms using IDs 4-6 and IDs 10-12. The results show that direct distillation did not enhance the performance and even led to a decrease in stability (RMSE, $a_1$). For the contrastive method, although we applied the constraint of Eq. \ref{ph} to the contrastive objects, directly contrasting the predicted noises would lead to network crashes (\textit{i.e.}, all predicted depth values are zeros, which are not included in Table \ref{xr2}). Therefore, we still maintain the noise constraint from Eq. \ref{ddimloss}, only taking the feature and image levels comparison, but there is only a minor improvement. The proposed trinity contrast is the only approach that significantly improves convergence on WeatherKITTI and demonstrates increased robustness and accuracy on real datasets.

\textbf{Effectiveness of Multi-Level Contrasts.}
\label{multi-xr}
The comparison of ablation with multi-level contrasts is presented in Table \ref{xr2} (IDs 2,3,6 and IDs 8,9,12). While the performance of different models is close in the WeatherKITTI dataset, in real-world weather situations, we notice that the results become more reliable as we gradually include the feature and image level trinity. The enhancements in these metrics, coupled with Fig. \ref{mult-vis}, apparently prove the effectiveness of the multi-level contrast.

\textbf{Compute Analysis.}
As shown in Table \ref{compute}, D4RD's diffusion process increases parameter count by 0.4\% and computational effort by 23.3\% compared to the baseline (traditional MDE). However, this is justified by a 17.5\% reduction in AbsRel error. The training process does incur higher memory and time expenses, but these can be mitigated by using multiple GPUs, and training is a one-time process. For inference, D4RD is slower due to iterative denoising, but several acceleration techniques have been successfully applied to solve these(see \cite{cvprdfs}'s GitHub). Additionally, the two-stage strategy is employed only during training, whereas the inference process remains end-to-end.

\vspace{-0.2cm}
\section{Conclusion}
In this paper, we enhance the stability and convergence of diffusion-based MDE and further introduce a novel robust depth estimation framework named D4RD, which integrates a customized contrastive learning scheme. This method combines the strengths of distillation with contrastive learning and repurposes the sampled noise in the diffusion forward process, creating a "trinity" contrastive mode, thereby improving the accuracy and robustness of noise prediction. Furthermore, we extend the noise level's trinity to the more general feature and image levels, constructing a multi-level trinity scheme to balance the robust perception pressure across different model components. Through extensive quantitative and qualitative experiments, we demonstrate the effectiveness of D4RD against various architectures and its superior performance over existing SoTA solutions.

\newpage
\begin{acks}
This work was supported by the National Natural Science Foundation of China (Nos. 62172032).
\end{acks}

\bibliographystyle{ACM-Reference-Format}
\balance
\bibliography{sample-base}

\newpage
\twocolumn[
\begin{center}
    {\LARGE \bfseries Supplementary Materials}
\end{center}
\vspace{0.5cm}
]
\section{Evaluation Metrics}
Similar to \cite{vit}, we employ the following  evaluation metrics in our experiments,
\begin{itemize}
    \item \textbf{AbsRel:}  $\ \frac{1}{|\mathcal{M}|} \sum_{d \in \mathcal{M}}\left|d-d_{gt}\right| / d_{gt}$;
\vspace{0.3em}
    \item \textbf{SqRel:}   $\frac{1}{|\mathcal{M}|} \sum_{d \in \mathcal{M}}\left\|d-d_{gt}\right\|^2 / d_{gt}$;
\vspace{0.3em}
    \item \textbf{RMSE}: $\sqrt{\frac{1}{|\mathcal{M}|} \sum_{d \in \mathcal{M}}\left\|d-d_{gt}\right\|^2} $;
\vspace{0.3em}
    \item \textbf{RMSElog}: $\sqrt{\frac{1}{|\mathcal{M}|} \sum_{d \in \mathcal{M}}\left\|\log (d)- \log (d_{gt})\right\|^2} $;
\vspace{0.3em}
    \item $\textbf{a}_{t}$: percentage of $d$ such that $\max(\frac{d}{d_{gt}},\frac{d_{gt}}{d}) < 1.25^t $ 
\end{itemize}
where $d_{gt}$ and $d$ denote the GT and estimated pixel depth, $\mathcal{M}$ is the valid mask set to $1e-3<d_{gt}<80$. Note that we also use the median scaling technique introduced by \cite{zhou} on $d$ to recover the absolute depth scale.

\vspace*{-0.5em}
\section{Formula symbol summary}
In this paper, to explain our method clearly, we use numerous symbols in the formulas and the meaning of each symbol can be found at its first appearance. However, for the readers' convenience, we once again list the symbol meanings in Table. \ref{symbol-tab}. Note that we only record meta symbols here and do not include compound symbols with superscripts and subscripts. The meaning of any symbol in this paper can be grasped through combinations. (e.g. $\textbf{d}_{S,clr}$ denote the depth prediction of the clear images from the student model).

\begin{table}[!ht]
\caption{\label{symbol-tab} The meanings of meta symbols. Because the same character can represent multiple meanings, please pay attention to font changes. }
\vspace*{-0.5em}
\resizebox{\linewidth}{!}{
\begin{tabular}{c|l}
\hline
\textbf{Symbol}&\textbf{Meaning}\\
\hline
\hline
a & The auxiliary image from adjacent frame \\
$\alpha$ & A transformation of $\beta$ \\
aug & The augmented image \\
$\beta$& Predefined variance noise schedule \\
\textbf{c} & The diffusion's condition \\
clr & The clear image \\
cst & Contrast (a learning method) \\
d & Pixel level depth\\
\textbf{d} & Map level depth prediction \\
D & Map level depth distribution (from GT) \\
dis & Distillation (a learning method)\\
e & The edges \\
$\epsilon$ & A noise that follows $\mathcal{N}$(0, I) distribution \\
f, feat & The depth feature \\
\multirow{2}{*}{$\mathcal{F}$} & A neural network, the concrete meaning is \\
                                &determined by its index \\
gt & The ground truth depth\\
I, img & The RGB images \\
L & The training loss \\
M & Adaptive distillation mask \\
$\mathcal{M}$ & The valid mask of GT depth\\
$\mathcal{N}$ & Gaussian sampling \\
nis & The noise \\
ph & Photometric consistency \\
S & Student model\\
T & Teacher model\\
$\mathcal{T}$ & Relative camera poses transformation \\
$\mathbb{T}$ & Overall diffusion training steps \\
t & The target image (The image to be estimated depth) \\
$\tau$ & The time step \\
$\mathcal{U}$ & Uniformly sampling \\
$\lambda$ & A constant, relative to adaptive mask\\
$\omega, \eta, \rho$ & Weights for various losses \\

\hline
\end{tabular}
}
\vspace*{-0.5em}
\end{table}
\vspace*{-0.5em}

\vspace*{-0.5em}
\section{The snowy dataset}

\begin{figure*}[ht]
\centering
\vspace*{-0.5em}
\includegraphics[width=0.9\textwidth]{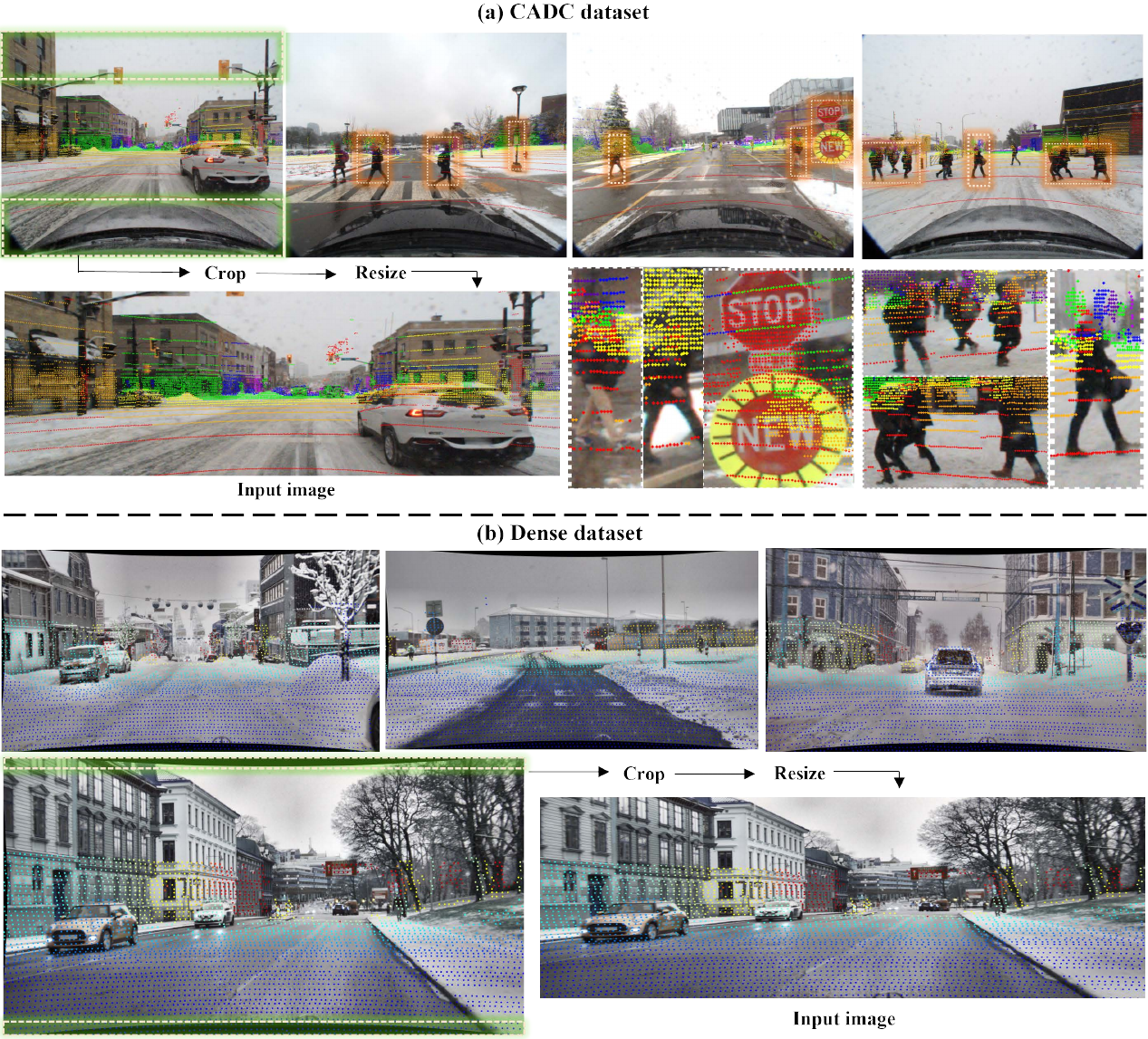}
\caption{\label{data-vis} The LiDAR points projection on the CADC dataset and Dense dataset. The cropped areas and prominent errors are highlighted with green and orange boxes, respectively. For the CADC dataset, the colors from near to far are red, orange, yellow, green, blue, and purple in order. And for Dense, it is blue, light blue, yellow, orange, and red.}
\vspace*{-1em}
\end{figure*}

In WeatherDepth\cite{wjy}, Wang et al. use the CADC dataset to validate their model's performance in snowy scenes. However, as shown in Fig. \ref{data-vis}(a), the LiDAR point clouds in this dataset have many errors on pedestrians and road signs. Furthermore, there is a vehicle bonnet at the bottom of each image, so Wang et al. have to crop them and the sky part, which greatly disrupts the depth clues brought by the vertical position. Hence, the experiments on this dataset can only give a rough result and we need a better depth-annotated snowy dataset.

In this work, we chose the Dense dataset. As depicted in Fig. \ref{data-vis}(b), the LiDAR points in Dense are more accurate and only require cropping a small portion of the image. All of these factors make the comparison result reasonable.

\vspace*{-0.5em}
\section{Additional Qualitative Results}

\begin{figure*}[ht]
\centering
\hspace*{-0.4em}
\includegraphics[width=1.02\textwidth]{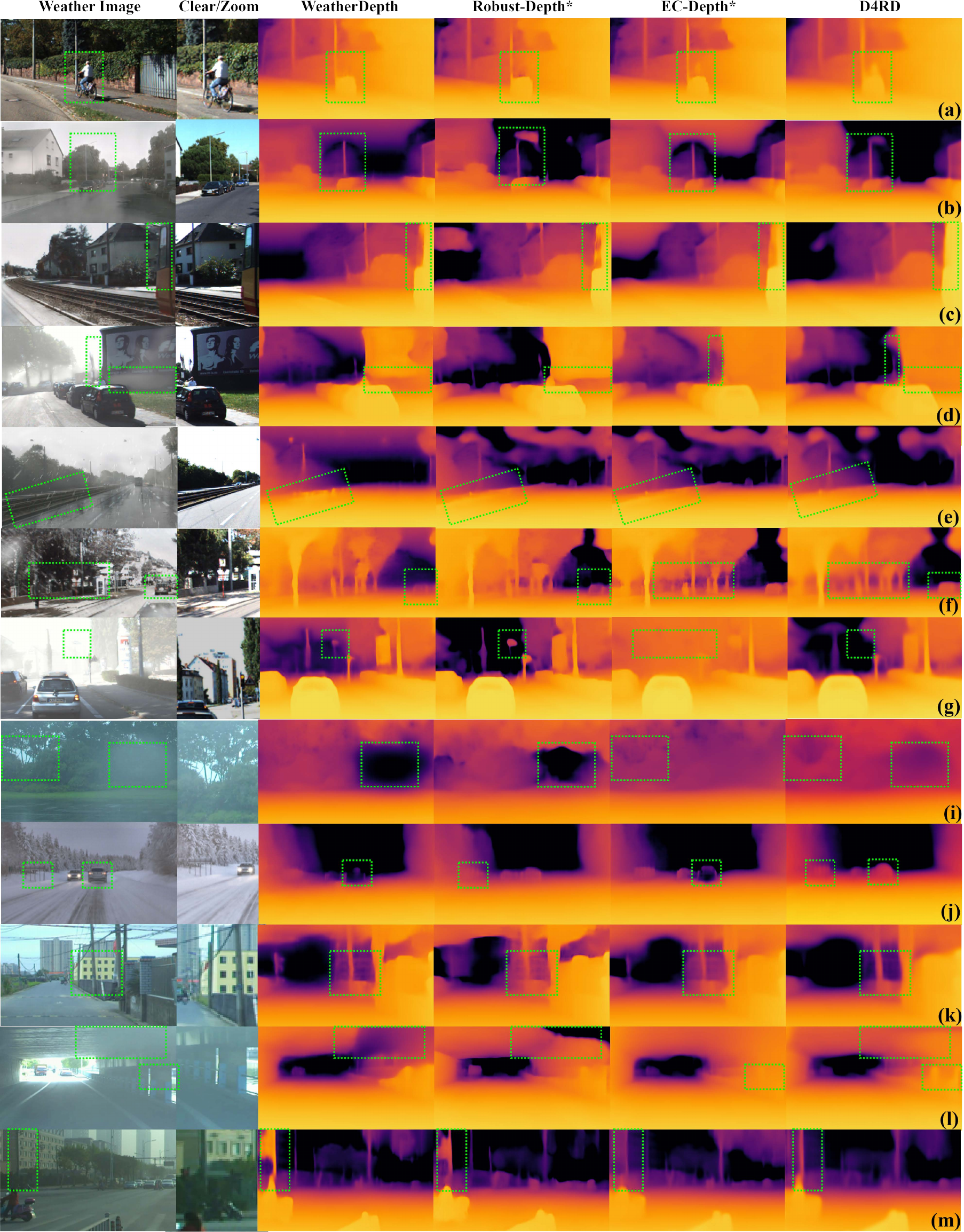}
\caption{\label{sup-vis} Extra visual comparisons on synthetic and real weather dataset. It is better viewed when zooming in. (a)-(g) are the WeatherKITTI dataset results and  (i)-(m) are the real weathers.}
\end{figure*}

To further vividly demonstrate the superiority of our method, as shown in Fig. \ref{sup-vis}, we present a visual comparison of each scene from both synthetic and real datasets. 
\textbf{(a)} is clear scene. The highlighted area in the dashed box exemplifies how pedestrians wearing white clothing are accurately distinguished from the background by D4RD.
\textbf{(b)} is synthetic light rainy scene. Despite being affected by camera blur, D4RD still recognizes the correct shape of the streetlights.
\textbf{(c)} is synthetic light snowy scene. D4RD stands out as the only RMDE model capable of not mistaking train windows for background depth.
\textbf{(d)} is synthetic light foggy scene. D4RD correctly recognizes that the wall is a plane, avoiding concave depth errors.
\textbf{(e)} is synthetic adverse rainy scene. The influence of railway tracks on depth should be negligible, but under the interference of raindrop particles, only D4RD exhibits correct depth structures.
\textbf{(f)} is synthetic adverse snowy scene. D4RD not only identifies the white car far away but also recognizes the almost invisible road sign (on the left side of the image).
\textbf{(g)} is synthetic adverse foggy scene. From the clear image, it can be seen that the object highlighted by the green box is a distant building roof, which should not be recorded in the depth map (given a maximum depth of 80m). Only D4RD can discern such details accurately.
\textbf{(i)} is the real rainy scene and a failure case for all RMDE models. Weatherdepth and Robust-Depth* identified lens blurring as infinity, while EC-Depth* did not recognize the sparse areas. Although D4RD addresses these issues to some extent, it remains flawed. This imperfection is likely attributed to insufficient training data available for this specific setting.
\textbf{(j)} is real snowy scene. Once again, the distant cars and fences are accurately identified by D4RD.
\textbf{(k)} is real cloudy scene. From the zoomed image, it can be seen that the building should be located behind a distant pole, which exceeds the range of 80m. However, D4RD can accurately identify this case.
\textbf{(l)} is from a real sunny scene, specifically focusing on the unique condition of the bridge cave. Affected by this out-of-distribution environment, the depth estimation performance of other RMDE models was significantly impacted. Conversely, D4RD demonstrates robustness and maintains accurate depth estimation.
\textbf{(m)} is real fog scene. From the zoomed image, it can be seen that the correct structural relationship of the highlighted area is `a pole in front of the distant building'. Other models either estimate the depth of the building too closely or estimate the depth of the poles too far. Only D4RD is competent for this job.

Based on the above discussion, we further prove the robustness and effectiveness of D4RD.

\end{document}